\documentclass{article}
\usepackage{ijcai20}

\usepackage{times}
\usepackage{soul}
\usepackage{url}
\usepackage[hidelinks]{hyperref}
\usepackage[utf8]{inputenc}
\usepackage[small]{caption}
\usepackage{graphicx}
\usepackage{amsmath}
\usepackage{amsthm}
\usepackage{booktabs}
\usepackage{algorithm}
\usepackage{algorithmic}
\usepackage{amssymb}

\usepackage{epstopdf}
\usepackage{booktabs}
\usepackage{makecell}
\usepackage{bm}
\usepackage{multirow}
\usepackage{subcaption}
\usepackage{enumitem}
\usepackage{ragged2e}
\usepackage{verbatim}

\usepackage{pdfpages}

\newcommand{\ie}{\emph{i.e., }}
\newcommand{\eg}{\emph{e.g., }}
\newcommand{\etal}{\emph{et al. }}
\newcommand{\etc}{\emph{etc. }}
\newcommand{\wrt}{\emph{w.r.t. }}

\title{Bilinear Graph Neural Network with Neighbor Interactions
\thanks{This work is supported by the National Natural Science Foundation of China (U19A2079, 61525206, 61972069, 61836007, 61832017). Fuli Feng is the corresponding author and contributes equally as Hongmin Zhu.}
}

\author{
Hongmin Zhu$^1$\and
Fuli Feng$^2$\and
Xiangnan He$^{1}$\and
Xiang Wang$^2$\\
Yan Li$^3$\and
Kai Zheng$^4$\And
Yongdong Zhang$^1$
\affiliations
$^1$University of Science and Technology of China\\
$^2$National University of Singapore\\
$^3$Beijing Kuaishou Technology Co., Ltd. Beijing, China\\
$^4$University of Electronic Science and Technology of China
\emails
\{zhuhm@mail., hexn@, zhyd73@\}ustc.edu.cn, \{dcsfeng, dcswxi\}@nus.edu.sg\\
liyan@kuaishou.com, zhengkai@uestc.edu.cn
}

\begin{document}

\maketitle
\begin{abstract} 
\textit{Graph Neural Network} (GNN) is a powerful model to learn representations and make predictions on graph data. Existing efforts on GNN have largely defined the graph convolution as a weighted sum of the features of the connected nodes to form the representation of the target node. Nevertheless, the operation of weighted sum assumes the neighbor nodes are independent of each other, and ignores the possible interactions between them. When such interactions exist, such as the co-occurrence of two neighbor nodes is a strong signal of the target node's characteristics, existing GNN models may fail to capture the signal. In this work, we argue the importance of modeling the interactions between neighbor nodes in GNN. We propose a new graph convolution operator, which augments the weighted sum with pairwise interactions of the representations of neighbor nodes. We term this framework as \textit{Bilinear Graph Neural Network} (BGNN), which improves GNN representation ability with bilinear interactions between neighbor nodes. In particular, we specify two BGNN models named BGCN and BGAT, based on the well-known GCN and GAT, respectively. Empirical results on three public benchmarks of semi-supervised node classification verify the effectiveness of BGNN --- BGCN (BGAT) outperforms GCN (GAT) by 1.6\% (1.5\%) in classification accuracy. 
Codes are available at: https://github.com/zhuhm1996/bgnn.
\end{abstract}
\section{Introduction}
GNN is a kind of neural networks that performs neural network operations over graph structure to learn node representations. Owing to the ability to learn more comprehensive node representations than the models that consider only node features~\cite{yang2016revisiting} or graph structure~\cite{perozzi2014deepwalk}, GNN has been a promising solution for a wide range of applications in social science~\cite{chen2018fastgcn}, 
computer vision~\cite{kampffmeyer2019rethinking}, and recommendation~\cite{wang2019neural,he2020lightgcn} \etc  To date, most graph convolution operations in GNNs are implemented as a linear aggregation (\ie weighted sum) over features of the neighbors of the target node~\cite{kipf2016semi}. Although it improves the representation of the target node, such linear aggregation assumes that the neighbor nodes are independent of each other, ignoring the possible interactions between them.

\begin{figure}[t]
    \centering
    \includegraphics[width=0.47\textwidth]{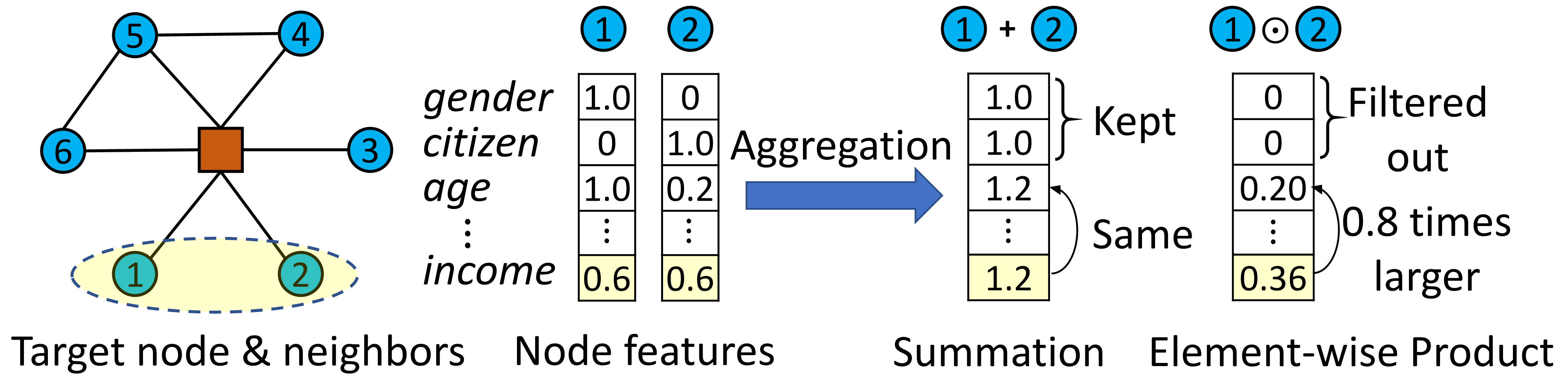}
    \caption{\textbf{Left}: A toy example of a target node and its neighbors in a transaction graph where nodes are described by a set of numerical features. \textbf{Right}: Results of aggregating the features of node 1 and 2 with summation and element-wise product operations, respectively.}
    \label{fig:BI}
\end{figure}

Under some circumstances, the interactions between neighbor nodes could be a strong signal that indicates the characteristics of the target node. Figure~\ref{fig:BI} (left) illustrates a toy example of a target node and its neighbors in a transaction graph, where edges denote money transfer relations and nodes are described by a set of features such as age and income. The interaction between node 1 and 2, which indicates that both have high incomes, could be a strong signal to estimate the credit rating of the target node (an intuition is that a customer who has close business relations with rich friends would have a higher chance to repay a loan). 
Explicitly modeling such interactions between neighbors highlights the common properties within the local structure, which could be rather helpful for the target node's representation. 
In Figure~\ref{fig:BI} (right), we show that the summation-based linear aggregator --- a common choice in existing GNNs --- fails to highlight the income feature. In contrast, by using a multiplication-based aggregator that captures node interactions, the signal latent in shared high incomes is highlighted, and as an auxiliary effect, some less useful features are zeroed out.

Nevertheless, it is non-trivial to encode such local node interactions in GNN. The difficulty mainly comes from two indispensable requirements of a feasible graph convolution operation: 1) \textit{permutation invariant}~\cite{xu2018powerful}, \ie the output should remain the same when the order of neighbor nodes is changed, so as to ensure the stability of a GNN; and 2) \textit{linear complexity}~\cite{kipf2016semi}, \ie the computational complexity should increase linearly with respect to the number of neighbors, so as to make a GNN scalable on large graphs. 
To this end, we take inspiration from neural factorization machines~\cite{he2017neural} to devise a new \textit{bilinear aggregator}, which explicitly expresses the interactions between every two nodes and aggregates all pair-wise interactions to enhance the target node's representation.

On this basis, we develop a new graph convolution operator which is equipped with both traditional linear aggregator and the newly proposed bilinear aggregator, and is proved to be permutation invariant. 
We name the new model as \textit{Bilinear Graph Neural Network} (BGNN), which is expected to learn more comprehensive representations by considering local node interactions. 
We devise two BGNN models, named BGCN and BGAT, which are equipped with the GCN and GAT linear aggregator, respectively. 
Taking semi-supervised node classification as an example task, we evaluate BGCN and BGAT on three benchmark datasets to validate their effectiveness. Specifically, BGCN and BGAT outperform GCN and GAT by 1.6\% and 1.5\%, respectively. More fine-grained analyses show that the improvements on sparsely connected nodes are more significant, demonstrating the strengths of the bilinear aggregator in modeling node interactions. 
The main contributions of this paper are summarized as:
\begin{itemize}[leftmargin=*]
    \item We propose BGNN, a simple yet effective GNN framework, which explicitly encodes the local node interactions to augment conventional linear aggregator.
    \item We prove that the proposed BGNN model has the properties of permutation invariant and linear computation complexity which are of importance for GNN models.
    \item We conduct extensive experiments on three public benchmarks of semi-supervised node classification, validating the effectiveness of the proposed BGNN models.
\end{itemize}
\section{Related Work}
GNN generalizes traditional convolutional neural networks from Euclidean space to graph domain. 
According to the format of the convolution operations, existing GNN models can be divided into two categories: spatial GNN and spectral GNN~\cite{zhang2018deep}. We separately review the two kinds of models, and refer the mathematical connection between them to~\cite{bronstein2017geometric}.

\paragraph{Spectral GNN.} Spectral GNN is defined as performing convolution operations in the Fourier domain with spectral node representations~\cite{bruna2013spectral,defferrard2016convolutional,kipf2016semi,liao2018lanczosnet,xu2018graph}. Bruna \etal \cite{bruna2013spectral} define the convolution over the eigenvectors of graph Laplacian which are viewed as the Fourier basis. 
Considering the high computational cost of the eigen-decomposition, research on spectral GNN has been focused on approximating the decomposition with different mathematical techniques~\cite{defferrard2016convolutional,kipf2016semi,liao2018lanczosnet,xu2018graph}. 
For instance, \cite{defferrard2016convolutional} introduce the Chebyshev polynomials with orders of $K$ to approximate the eigen-decomposition. 
In~\cite{kipf2016semi}, Kipf and Welling simplify this model by limiting $K = 1$ and approximating the largest eigenvalue of Laplacian matrix by $2$. 
In addition, Liao \etal \cite{liao2018lanczosnet} employ the Lanczos algorithm to perform a low-rank approximation of the graph Laplacian. 
Recently, Wavelet transform is introduced to spectral GNN to discard the eigen-decomposition~\cite{xu2018graph}.   
However, spectral GNN models are hard to be applied on large graphs such as social networks. This is because the convolution operations are required to be performed over the whole graph, posing unaffordable memory cost and incapacitating the widely applied batch training. 
\paragraph{Spatial GNN.} 
Spatial GNN instead performs convolution operations directly over the graph structure by aggregating the features from spatially close neighbors to a target node~\cite{atwood2016diffusion,hamilton2017inductive,kipf2016semi,velivckovic2017graph,xu2018representation,xinyi2018capsule,velickovic2018deep,xu2018powerful,feng2019temporal}. 
This line of research is mainly focused on developing aggregation methods from different perspectives. For instance, 
Kipf and Welling~\cite{kipf2016semi} propose to use a linear aggregator (\ie weighted sum) that uses the reverse of node degree as the coefficient. 
To improve the representation performance, neural attention mechanism is introduced to learn the coefficients~\cite{velivckovic2017graph}. 
In addition to aggregating information from directly connected neighbors, augmented aggregators also account for multi-hop neighbors~\cite{atwood2016diffusion,xu2018representation}. 
Moreover, non-linear aggregators are also employed in spatial GNNs such as max pooling~\cite{hamilton2017inductive}, capsule~\cite{velickovic2018deep}, and Long Short-Term Memory (LSTM)~\cite{hamilton2017inductive}. Furthermore, spatial GNN is extended to graphs with both static and temporal neighbors structure~\cite{park2019exploiting} and representations in hyperbolic space~\cite{chami2019hyperbolic}.

However, most existing aggregators (both linear and non-linear ones) forgo the importance of the interactions among neighbors. As built upon the summation operation, by nature, the linear aggregators assume that neighbors are independent. 
Most of the non-linear ones are focused on the property of neighbors at set level (\ie all neighbors), \eg the "skeleton" of the neighbors~\cite{xu2018powerful}. Taking one neighbor as the input of a time-step, LSTM-based aggregator could capture 
sequential dependency, which might include 
node interactions. However, it requires a predefined order on neighbor, violating permutation invariant and typically showing weak performance~\cite{hamilton2017inductive}. Our work is different from those aggregators
in that we explicitly consider pairwise node interactions in a neat and systematic way.   

\section{Bilinear Graph Neural Network}
\label{section3}

\paragraph{Preliminaries.}
Let $G=(\textbf{A}\in \{0,1\}^{N\times N}, \textbf{X}\in \mathbb{R}^{N\times F})$ be the graph of interest, where $\textbf{A}$ is the binary adjacency matrix where an element $A_{vi} = 1$ means that an edge exists between node $v$ and $i$, and $\textbf{X}$ is the original feature matrix for nodes that describes each node with a vector of size $F$ (a row).  
We denote the neighbors of node $v$ as $\mathcal{\bm{N}}(v)=\{i | A_{vi}=1\}$ which stores all nodes that have an edge with $v$, and denote the extended neighbors of node $v$ as $\tilde{\mathcal{\bm{N}}}(v)=\{v\}\cup  \mathcal{\bm{N}}(v)$ which contains the node $v$ itself. 
For convenience, we use $d_v$ to denote the degree of node $v$, \ie $d_v = |\mathcal{{N}}(v)|$, and accordingly $\tilde{d}_v=|\tilde{\mathcal{\bm{N}}}(v)|=d_v+1$. 
The model objective is to learn a representation vector $\textbf{h}_v\in \mathbb{R}^D$ for each node $v$, such that its characteristics are properly encoded. For example, the label of node $v$ can be directly predicted as a function output $y_{v} = f(\textbf{h}_v)$, without the need of looking into the graph structure and original node features in $G$.

The spatial GNN~\cite{velivckovic2017graph} achieves this goal by recursively aggregating the features from neighbors:
\begin{equation}
\footnotesize{
\begin{split}
\textbf{h}^{(k)}_v = AGG\big(\{\textbf{h}^{(k-1)}_i\}_{i\in \tilde{\mathcal{\bm{N}}}(v)} \big) = \sum_{i\in \tilde{\mathcal{\bm{N}}}(v)} a_{vi} \textbf{h}^{(k-1)}_i \textbf{W}^{(k)},
\end{split}
}
\end{equation}
where $\textbf{h}^{(k)}_v$ denotes the representation of target node $v$ at the $k$-th layer/iteration, $\textbf{W}^{(k)}$ is the weight matrix (model parameter) to do feature transformation at the $k$-th layer, and the initial feature representation $\textbf{h}^{(0)}_v$ can be obtained from the original feature matrix $\textbf{X}$. 

The $AGG$ function is typically implemented as a weighted sum with $a_{vi}$ as the weight of neighbor $i$. In GCN~\cite{kipf2016semi}, $a_{vi}$ is defined as $1 / \sqrt{\tilde{d}_v\tilde{d}_i}$, which is grounded on the Laplacian theories. The recent advance on graph attention network (GAT)~\cite{velivckovic2017graph} learns $a_{vi}$ from data, which has the potential to lead better performance than pre-defined choices. 
However, a limitation of such weighted sum is that no interactions between neighbor representations are modeled. Although using more powerful feature transformation function such as multi-layer perceptron (MLP)~\cite{xu2018powerful} can alleviate the problem, the process is rather implicit and ineffective. 
An empirical evidence is from \cite{beutel2018latent}, which shows that MLP is inefficient in capturing the multiplication relations between input features. In this work, we propose to explicitly inject the multiplication-based node interactions into $AGG$ function.  

\subsection{Bilinear Aggregator}
\label{section3.1}
As demonstrated in Figure~\ref{fig:BI}, the multiplication between two vectors is an effective manner to model the interactions --- emphasizing common properties and diluting discrepant information. 
Inspired by factorization machines (FMs)~\cite{rendle2010factorization,he2017neural} that have been intensively used to learn the interactions among categorical variables, we propose a bilinear aggregator which is suitable for modeling the neighbor interactions in local structure:
\begin{equation}\label{eq:bi-agg}
\footnotesize{
\begin{split}
&BA \big(\{ \textbf{h}_i \}_{i \in \tilde{\mathcal{\bm{N}}}(v)} \big) = \frac{1}{b_v} \sum_{i\in \tilde{\mathcal{\bm{N}}}(v)} \sum_{j\in \tilde{\mathcal{\bm{N}}}(v)\& i\textless j} \textbf{h}_{i}\textbf{W}\odot \textbf{h}_{j}\textbf{W},
\end{split}
}
\end{equation}
where $\odot$ is element-wise product; $v$ is the target node to obtain representation for; $i$ and $j$ are node index from the extended neighbors $\tilde{\mathcal{\bm{N}}}(v)$ ---  they are constrained to be different to avoid self-interactions that are meaningless and may even introduce extra noises. $b_{v}=\frac{1}{2} \tilde{d}_v(\tilde{d}_v-1)$ denotes the number of interactions for the target node $v$, which normalizes the obtained representation to remove the bias of node degree. 
It is worth noting that we take the target node itself into account and aggregate information from extended neighbors, which although looks same as GNN, but for different reasons. In GNN, accounting for the target node is to retain its information during layer-wise aggregation, working like the
residual learning~\cite{he2016deep}. While in BGNN, our consideration is that 
the interactions between the target node and its neighbors may also carry useful signal. For example, for sparse nodes that have only one neighbor, the interaction between neighbors does not exist, and the interaction between the target and neighbor nodes can be particularly helpful. 

\paragraph{Time Complexity Analysis.}
At the first sight, the bilinear aggregator considers all pairwise interactions between neighbors (including the target node), thus may have a quadratic time complexity \textit{w.r.t.} the neighbor count, being higher than the weighted sum. Nevertheless, through a mathematical reformulation similar to that one used in FM, we can compute the aggregator in linear time --- $\mathcal{O}(|\tilde{\mathcal{\bm{N}}}(v)|)$ --- the same complexity as weighted sum. To show this, we rewrite Equation (\ref{eq:bi-agg}) in its equivalent form as:  
\begin{equation}\label{eq:reduce}
\footnotesize{
\begin{aligned}
\begin{split}
&BA \big(\{ \textbf{h}_i \}_{i \in \tilde{\mathcal{\bm{N}}}(v)} \big) = 
	\frac{1}{2 b_v} \bigg( \sum_{i\in \tilde{\mathcal{\bm{N}}}(v)} \sum_{j\in \tilde{\mathcal{\bm{N}}}(v)} \textbf{s}_i\odot \textbf{s}_j \\&-
	\sum_{i \in \tilde{\mathcal{\bm{N}}}(v)} \textbf{s}_i \odot \textbf{s}_i \bigg) 
= \frac{1}{2 b_v}\bigg(\Big(
\underbrace{\sum_{i\in \tilde{\mathcal{\bm{N}}}(v)}\textbf{s}_i}_{\mathcal{O}(|\tilde{\mathcal{\bm{N}}}(v)|)}\Big)^{2}-
\underbrace{\sum_{i\in \tilde{\mathcal{\bm{N}}}(v)}\textbf{s}_i^{2}}_{\mathcal{O}(|\tilde{\mathcal{\bm{N}}}(v)|)}
\bigg),
\end{split}
\end{aligned}
}
\end{equation}
where $\textbf{s}_i = \textbf{h}_i \textbf{W} \in \mathbb{R}^D$.
As can be seen, through mathematical reformulation, we can reduce the sum over pairwise element-wise products to the minus of two terms, where each term is a weighted sum of neighbor representations (or their squares) and can be computed in $\mathcal{O}(|\tilde{\mathcal{\bm{N}}}(v)|)$ time. Note that multiplying weight matrix $\textbf{W}$ is a standard operation in aggregator thus its time cost is omitted for brevity. 

\paragraph{Proof of Permutation Invariant.}  
This property is intuitive to understand from the reduced Equation~(\ref{eq:reduce}): when changing the order of input vectors, the sum of inputs (the first term) and the sum of the squares of inputs (the second term) are not changed. Thus the output is unchanged and the permutation invariant property is satisfied. 
To provide a rigorously proof, we give the matrix form of the bilinear aggregator, which also facilitates the matrix-wise implementation of BGNN. 
The matrix form of the bilinear aggregator is:
\begin{equation}\label{eq:matrix}\footnotesize
BA(\textbf{H},\textbf{A}) = 
	\frac{1}{2} \textbf{B}^{-1} 
	\bigg(
		\Big(
			\tilde{\textbf{A}} \textbf{HW}
		\Big)^{2} -
		\tilde{\textbf{A}} 
		\Big(
			\textbf{HW}
		\Big)^{2} 
	\bigg),
\end{equation}
where $\textbf{H}\in \mathbb{R}^{N\times D}$ stores the representation vectors $\textbf{h}$ for all nodes, 
$\tilde{\textbf{A}}=\textbf{A}+\textbf{I}$ is the adjacency matrix of the graph with self-loop added on each node ($\textbf{I}\in \mathbb{R}^{N\times N}$ is an identity matrix), $\textbf{B}$ is a diagonal matrix with each element $B_{vv}=b_v$, and $(\cdot)^{2}$ denotes the element-wise product of two matrices.

Let $\textbf{P}\in \mathbb{R}^{N\times N}$ be any permutation matrix that satisfies (1) $\textbf{P}^T \textbf{P} = \textbf{I}$, and (2) for any matrix $\textbf{M}$ if $\textbf{P}\textbf{M}$ exists, then $\textbf{P}\textbf{M}\odot \textbf{P}\textbf{M}=\textbf{P}(\textbf{M}\odot \textbf{M})$ satisfies. 
When we apply the permutation $\textbf{P}$ on the nodes, $\textbf{H}$ changes to $\textbf{PH}$, 
$\tilde{\textbf{A}}$ changes to $\textbf{P}\tilde{\textbf{A}}\textbf{P}^{T}$ and $\textbf{B}$ changes to $\textbf{P}\textbf{B}\textbf{P}^{T}$, which leads to:
\begin{equation*}\footnotesize
\begin{aligned}
\begin{split}
&BA (\textbf{P}\textbf{H},\textbf{P}\textbf{A}\textbf{P}^{T})
=\frac{1}{2}\textbf{P}\textbf{B}^{-1}\textbf{P}^{T}\bigg(\textbf{P}\Big(\tilde{\textbf{A}}\textbf{HW}\Big)^{2}-\textbf{P}\tilde{\textbf{A}}\Big(\textbf{HW}\Big)^{2} \bigg) \\
&=\frac{1}{2}\textbf{P}\textbf{B}^{-1}\textbf{P}^{T}\textbf{P}\bigg(\Big(\tilde{\textbf{A}}\textbf{HW}\Big)^{2}-\tilde{\textbf{A}}\Big(\textbf{HW}\Big)^{2} \bigg) = \textbf{P}\cdot BA (\textbf{H},\textbf{A}),
\end{split}
\end{aligned}
\end{equation*}
which indicates the permutation invariant property.

\subsection{BGNN Model}
\begin{figure}[t]
	\centering
	\hspace{6pt}
	\begin{subfigure}{.28\linewidth}
		\centering
		\includegraphics[width=\textwidth]{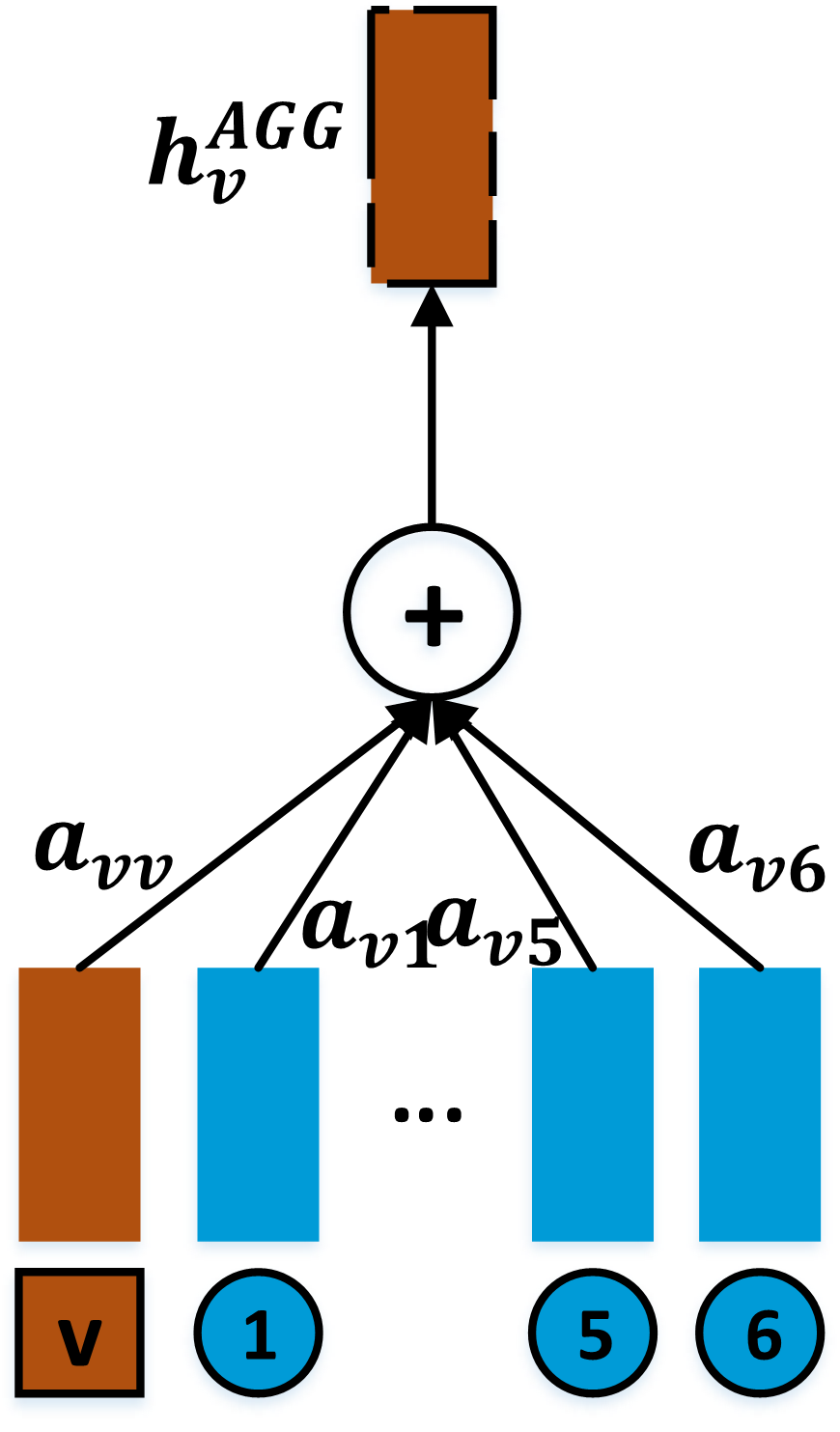}
		\caption{AGG}
		\label{fig2:a}
	\end{subfigure}%
	\hspace{6pt}
	\begin{subfigure}{.28\linewidth}
		\centering
		\includegraphics[width=\textwidth]{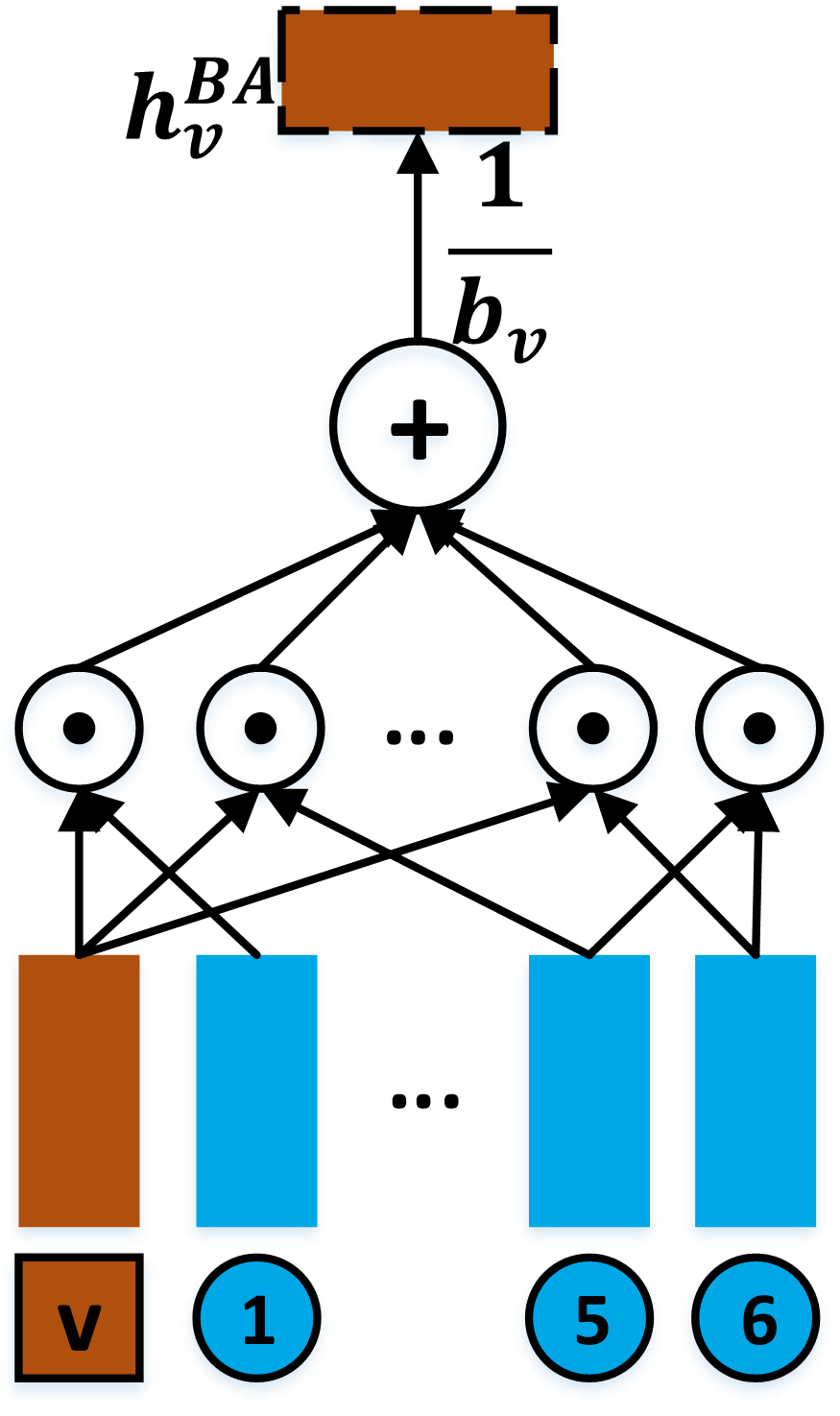}
		\caption{BA}
		\label{fig2:b}
	\end{subfigure}%
	\hspace{6pt}
	\begin{subfigure}{.28\linewidth}
		\centering
		\includegraphics[width=\textwidth]{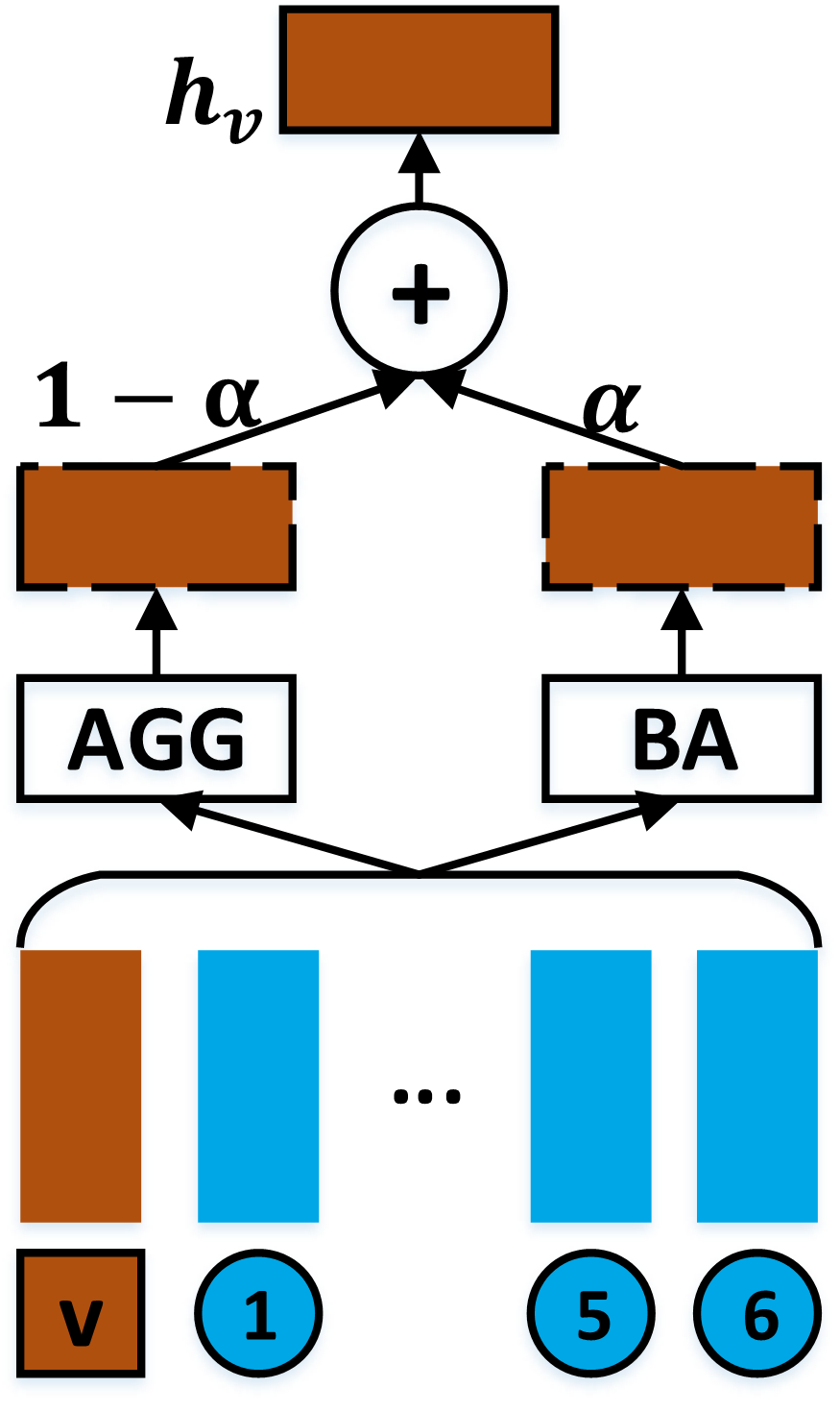}
		\caption{BGNN}
		\label{fig2:c}
	\end{subfigure}%
	\hfill
	\caption{An illustration of the traditional linear aggregator in GNN (a), bilinear aggregator (b), and BGNN aggregator (c).}
	\label{fig:framework}
\end{figure}

We now describe the proposed BGNN model. As the bilinear aggregator emphasizes node interactions and encodes different signal with the weighted sum aggregator, we combine them to build a more expressive graph convolutional network
We adopt a simple linear combination scheme, defining a new graph convolution operator as:
{\footnotesize
\begin{align}\label{eq:bgcn_layer}
\textbf{H}^{(k)} &= BGNN \Big( \textbf{H}^{(k - 1)}, \textbf{A} \Big) \\ \notag
&= 
(1 - \alpha)\cdot AGG 
\bigg(
	\textbf{H}^{(k-1)}, \textbf{A}
\bigg) + \alpha \cdot BA
\bigg(
	\textbf{H}^{(k-1)}, \textbf{A}
\bigg),
\end{align}
}
where $\textbf{H}^{(k)}$ stores the node representations at the $k$-th layer (encoded $k$-hop neighbors). $\alpha$ is a hyper-parameter to trade-off the strengths of traditional GNN aggregator and our proposed bilinear aggregator. Figure~\ref{fig:framework} illustrates the model framework. 

Since both $AGG$ and $BA$ are permutation invariant, it is trivial to find that this graph convolution operator is also permutation invariant. 
When $\alpha$ sets to 0, no node interaction is considered and BGNN degrades to GNN; when $\alpha$ sets to 1, BGNN only uses the bilinear aggregator to process the information from the neighbors. Our empirical studies show that an intermediate value between 0 and 1 usually leads to better performance, verifying the efficacy of modeling node interactions, and the optimal setting varies on different datasets. 

\paragraph{Multi-layer BGNN.} Traditional GNN models~\cite{kipf2016semi,velivckovic2017graph,xu2018powerful} encode information from multi-hop neighbors in a recursive manner by stacking multiple aggregators. For example, the 2-layer GNN model is formalized as,
\begin{equation}
\footnotesize{
	GNN_2(\textbf{X}, \textbf{A}) = \underbrace{AGG}_{\text{2nd layer}} \bigg( \sigma \Big( \underbrace{AGG}_{\text{1st layer}}(\textbf{X}, \textbf{A})\Big), \textbf{A} \bigg),
}
\end{equation}
where $\sigma$ is a non-linear activation function.   
Similarly, we can devise a 2-layer BGNN model in the same recursive manner:
\begin{equation}
\footnotesize{ 
\underbrace{BGNN}_{\text{2nd layer}} \bigg( \sigma \Big( \underbrace{BGNN}_{\text{1st layer}}(\textbf{X}, \textbf{A})\Big), \textbf{A} \bigg).
}
\end{equation}
However, such a straightforward multi-layer extension involves unexpected higher-order interactions. 
In the two-layer case, the second-layer representation will include partial 4th-order interactions among the two-hop neighbors, which are hard to interpret and unreasonable.  
When extending BGNN to multiple layers, saying $K$ layers, we still hope to capture pairwise interactions, but between the $K$-hop neighbors. To this end, instead of directly stacking $BGNN$ layers, we define the 2-layer $BGNN$ model as: 
{\footnotesize
\begin{align}
\label{eq:bgcn_2layer}
BGNN_2(\textbf{X}, \textbf{A}) &= (
	1 - \alpha
) \cdot GNN_2(\textbf{X}, \textbf{A}) \\ \notag
&+ \alpha [
	(1 - \beta) \cdot BA(\textbf{X}, \textbf{A}) + 
	\beta \cdot BA(\textbf{X}, \textbf{A}^{(2)}) ],
\end{align}
}
where $\textbf{A}^{(2)} = binarize(\textbf{A} \textbf{A})$ stores the 2-hop connectivities of the graph. $binarize$ is an entry-wise operation that transforms non-zero entries to 1.
As such, a non-zero entry $(v,i)$ in $\textbf{A}^{(2)}$ means node $v$ can reach node $i$ within two hops. 
$\beta$ is a hyper-parameter to trade-off the strengths of bilinear interactions within 1-hop neighbors and 2-hop neighbors. 

Following the same principle, we define the $K$-layer BGNN as: 
\begin{equation}\label{eq:bgcn_klayer}
\footnotesize{
	\begin{split}
	&BGNN_K(\textbf{X}, \textbf{A}) = (
		1 - \alpha
	) \cdot GNN_K(\textbf{X}, \textbf{A}) \\&+ 
	\alpha \cdot 
	\Big(
		\sum_{k=1}^{K} \beta_k \cdot BA(\textbf{X}, \textbf{A}^{(k)})
	\Big),\text{    }s.t., \sum_{k=1}^{K} \beta_k  = 1,
	\end{split}
}
\end{equation}
where $\textbf{A}^{(k)} = binarize\Big(\underbrace{\textbf{A} \cdots \textbf{A}}_{k \text{ times}}\Big)$ denotes the adjacency matrix of $k$-hop connectivities, and $GNN_K$ denotes normal $K$-layer GNN that can be defined recursively such as a $K$-layer GCN or GAT. The time complexity of a $K$-layer BGNN is determined by the number of non-zero entries in $\textbf{A}^{(K)}$. To reduce the actual complexity, one can follow the sampling strategy in GraphSage~\cite{hamilton2017inductive}, sampling a portion of high-hop neighbors rather than using all neighbors. 

\paragraph{Model Training.}
BGNN is a differentiable model, thus it can be end-to-end optimized on any differential loss with gradient descent. In this work, we focus on the semi-supervised node classification task, optimizing BGNN with the cross-entropy loss on labeled nodes (same setting as the GCN work~\cite{kipf2016semi} for a fair comparison). As the experimented data is not large, we implement the layer-wise graph convolution in its matrix form, leaving the batch implementation and neighbor sampling which can scale to large graphs as future work. 

\section{Experiments}
\label{sec:exp}
\label{section4.1}
\begin{table*}[]
	\centering
	\resizebox{0.71\textwidth}{!}{%
		\begin{tabular}{l|ccc|c|ccc|c}
			\hline
			\multirow{2}{*}{Model} & \multicolumn{3}{c|}{1-layer} &
			\multirow{2}{*}{RI} &
			\multicolumn{3}{c|}{2-layer} & \multirow{2}{*}{RI}\\ 
			& Pubmed & Cora & Citeseer & & Pubmed & Cora & Citeseer & \\ \hline \hline
			SemiEmb & - & - & - & - & $71.1$ & $59.0$ & $59.6$ & $26.4\%$\\ 
			DeepWalk & - & - & - & - & $65.3$ & $67.2$ & $43.2$ & $39.6\%$\\ 
			Planetoid & - & - & - & - & $77.2$& $75.7$& $64.7$  & $9.7\%$\\ 
			GCN & $76.9\pm 0.2$& $76.8\pm 0.2$& $69.1\pm 0.1$ & $2.7\%$ & $79.0$& $81.5$& $70.3$  & $3.2\%$\\ 
			GAT & $77.3\pm 0.4$& $78.3\pm 0.6$& $69.7\pm 1.1$ & $1.6\%$ & $79.0\pm 0.3$ & $83.0\pm 0.7$ & $72.5\pm 0.7$ & $1.5\%$\\ 
			GIN & $76.5\pm 0.1$& $77.5\pm 0.0$& $67.3\pm 0.9$  & $3.5\%$ & $78.5\pm 0.2$& $79.7\pm 0.8$& $69.4\pm 0.6$  & $4.6\%$\\
			\hline \hline
			BGCN-A & $77.7 \pm 0.2$& $79.0 \pm 0.2$& $70.1 \pm 0.0$& - & $79.1 \pm 0.2$ & $80.0 \pm 0.6$& $71.3 \pm 0.3$  & - \\ 
			BGCN-T & \textbf{78.0 $\pm$ 0.2}& $78.7\pm 0.2$& $70.6 \pm 0.1$ & - & $79.4 \pm 0.1$ & $82.0 \pm 0.1$& $71.9 \pm 0.0$  & - \\ \hline \hline
			BGAT-A & $77.6 \pm 0.3$& $78.6 \pm 1.2$& $71.0 \pm 1.4$& - & $79.1 \pm 0.4$ & $82.9 \pm 0.9$& $73.2 \pm 0.7$  & - \\
			BGAT-T & $77.8 \pm 0.2$& \textbf{79.6 $\pm$ 0.6}& \textbf{71.4 $\pm$ 1.3} & - & \textbf{79.8 $\pm$ 0.3} & \textbf{84.2 $\pm$ 0.4}& \textbf{74.0 $\pm$ 0.3}  & - \\ \hline
		\end{tabular}%
	}
	\caption{Performance of the compared methods on the three datasets \wrt prediction accuracy (mean of 10 different runs). The performance of GCN (2-layer) and GAT (2-layer) are copied from their original papers. RI means the average relative improvement across datasets achieved by BGAT-T. We omit the models with more layers for the consideration of over-smoothing issue~\protect\cite{li2018deeper}.
	}
	\label{tab:perf_comp}
\end{table*}
\paragraph{Datasets.}
Following previous works~\cite{sen2008collective,yang2016revisiting,velivckovic2017graph}, we utilize three benchmark datasets of citation network---Pubmed, Cora and Citeseer~\cite{sen2008collective}. In these datasets, nodes and edges represent documents and citation relations between documents, respectively. Each node is represented by the bag-of-words features extracted from the content of the document. Each node has a label with one-hot encoding of the document category. We employ the same data split in previous works~\cite{kipf2016semi,yang2016revisiting,velivckovic2017graph}. That is, 20 labeled nodes per class are used for training. 500 nodes and 1000 nodes are used as validation set and test set, respectively. Note that the train process can use all of the nodes' features. For this data split, we report the average test accuracy over ten different random initializations. To save space, we refer~\cite{kipf2016semi} for the detailed statistics of the three datasets.

\paragraph{Compared Methods.}
We compare against the strong baselines mainly in two categories: \textit{network embedding} and \textit{GNN}. We select three widely used network embedding approaches: graph regularization-based network embedding (SemiEmb)~\cite{weston2012deep} and skip-gram-based graph embedding (DeepWalk~\cite{perozzi2014deepwalk} and Planetoid~\cite{yang2016revisiting}). For GNNs, we select GCN~\cite{kipf2016semi}, GAT~\cite{velivckovic2017graph} and Graph Isomorphism Network (GIN)~\cite{xu2018powerful}.

We devise two BGNNs which implement the $AGG$ function as GCN and GAT, respectively. For each BGNN, we compare two variants with different scopes of the bilinear interactions: 1) BGCN-A and BGAT-A which consider all nodes within the $k$-hop neighbourhood, including the target node in the bilinear interaction.
2) BGCN-T and BGAT-T, which consider the interactions between the target node and the neighbor nodes within its $k$-hop neighbourhood.

\paragraph{Parameter Settings.}
We closely follow the GCN work~\cite{kipf2016semi} to set the hyper-parameters of SemiEmb, DeepWalk, and Planetoid. We perform grid-search to select the optimal values for hyper-parameters of the remaining methods, including the dropout rate, the weight for $l_2$-norm ($\lambda$), the $\beta$ trade-off the aggregated information from multi-hop nodes, and the $\alpha$ that balances the linear aggregator and bilinear aggregator. The dropout rates, $\lambda$, $\beta$ and $\alpha$ are selected within $[0, 0.2, 0.4, 0.6]$, $[0, 1e\textbf{-}4, 5e\textbf{-}4, 1e\textbf{-}3]$, $[0, 0.1, 0.3,\cdots, 0.9, 1]$ and $[0, 0.1, 0.3,\cdots, 0.9, 1]$, respectively. All BGNN-based models are trained for 2,000 epochs with an early stopping strategy based on both convergence behavior and accuracy of the validation set. 

\subsection{Performance Comparison}
\label{section4.2}
Table~\ref{tab:perf_comp} shows the performance of the compared methods on the three datasets \wrt prediction accuracy on the data split exactly same as in~\cite{kipf2016semi}. From the table, we have the following observations:
\begin{itemize}[leftmargin=*]
	\item In all cases, the proposed BGNN models achieves the best performance with average improvements over the baselines larger than 1.5\%. The results validate the effectiveness of BGNN which is attributed to incorporating the pairwise interactions between the nodes in the local structure (\ie the ego network of the target node) when performing graph convolution.
	\item On average, BGAT (BGCN) outperforms vanilla GAT (GCN) by 1.5\% (1.6\%). These results further indicate the benefit of considering the interaction between neighbor nodes, which could augment the representation of a target node, facilitating its classification. Furthermore, the improvements of BGAT and BGCN in the 1-layer and 2-layer settings are close, which indicates that the interactions between both 1-hop neighbors and 2-hop neighbors are helpful for the representation of a target node.
	\item BGNN models, which have different scopes of the bilinear interactions, achieve different performance across datasets. In most cases, BGAT-T (BGCN-T) achieves performance better than BGAT-A (BGCN-A), signifying the importance of interactions with the target node.
	\item Among the baselines, GCN models perform better than embedding-based methods, indicating the effectiveness of graph convolution operation in learning node representations. GAT models perform better than GCN models. These results are consistent with findings in previous works~\cite{kipf2016semi,yang2016revisiting,velivckovic2017graph}.
\end{itemize}
\begin{table}
	\centering
	\resizebox{0.37\textwidth}{!}{%
    \begin{tabular}{cc|ccc}
    	\hline
    	Split & Model&Pubmed &Cora &Citeseer \\
    	\hline\hline
    	\multirow{2}{*}{Random} & GCN&$77.0\pm 1.3$ &$79.7\pm 1.2$ &$70.8\pm 0.9$ \\
    	& BGCN-T&\textbf{77.9 $\pm$ 1.1} &\textbf{80.3 $\pm$ 1.1} &\textbf{71.6 $\pm$ 1.1} \\\hline
    	\multirow{2}{*}{Fixed} & GCN & $79.0$& $81.5$& $70.3$ \\
    	& BGCN-T & \textbf{79.4 $\pm$ 0.1} & \textbf{82.0 $\pm$ 0.1}& \textbf{71.9 $\pm$ 0.0} \\
    	\hline
    \end{tabular}
    }
    \caption{Test accuracy of 2-layer GCN and BGCN-T on the three datasets with random data splits.}
    \label{tab:random split}
\end{table}
As reported in~\cite{kipf2016semi} (Table 2), the performance of GCN on random data splits is significantly worse than the fixed data split. As such, following~\cite{wu2019simplifying}, we also test the methods on 10 random splits of the training set while keeping the validation and test sets unchanged. 
Table~\ref{tab:random split} shows the performance of BGCN-T and GCN over random splits. To save space, we omit the results of BGCN-A and BGAT-based models which show similar trends. As can be seen, BGCN-T still outperforms GCN with high significant level ($<5\%$), which further validates the effectiveness of the proposed model. However, the performance of both BGCN-T and GCN suffers from random data split as compared to the fixed data split. This result is consistent with previous work~\cite{wu2019simplifying} and reasonable since the hyper-parameters are tuned on the fixed data split.
\begin{figure}[t]
	\centering
	\begin{subfigure}{.43\linewidth}
		\centering
		\includegraphics[width=\textwidth]{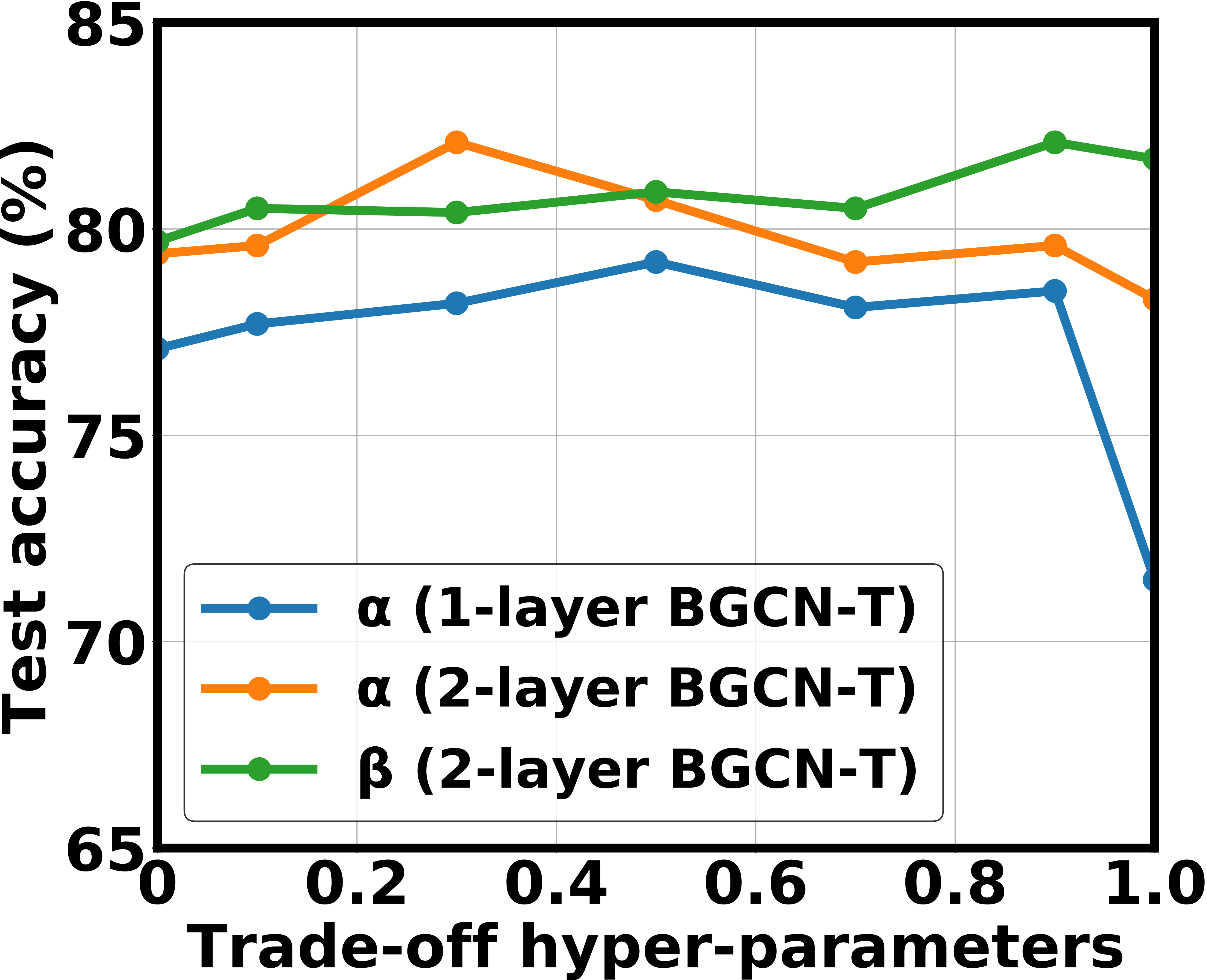}
		\caption{Cora}
		\label{fig3:a}
	\end{subfigure}%
	\begin{subfigure}{.43\linewidth}
		\centering
		\includegraphics[width=\textwidth]{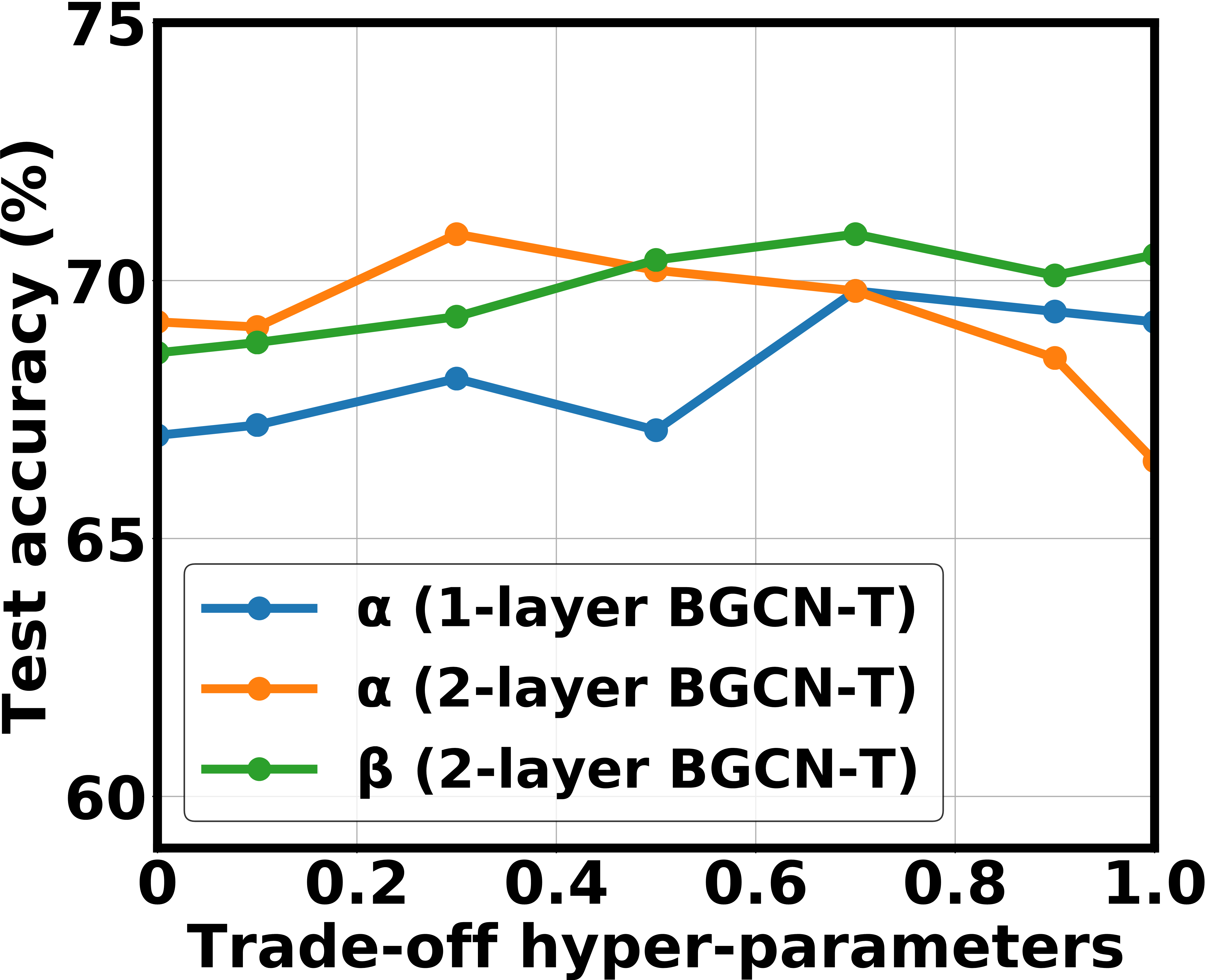}
		\caption{Citeseer}
		\label{fig3:b}
	\end{subfigure}%
	\caption{Impacts of trade-off hyper-parameters.}
	\label{fig:perf_alpha_beta}
\end{figure}

\subsection{Study of BGNN}
\paragraph{Impacts of Bilinear Aggregator.}
As the BA is at the core of BGNN, we first investigate its impacts on the performance by varying the value of $\alpha$.
Note that larger $\alpha$ means more contributions from the BA;
BGNN will downgrade to vanilla GNN with only the linear aggregator by setting $\alpha=0$, while being fully dependent on the BA by setting $\alpha=1$.
Figures~\ref{fig3:a} and~\ref{fig3:b} show the performance of BGCN-T with 1-layer and 2-layer on Cora and Citeseer datasets, respectively. We omit the results of other BGCN-based and BGAT-based models and results on Pubmed for saving space, which show similar trends. 
We have the following observations: 1) Under the two settings (1-layer and 2-layer), the performance of BGCN-T varies in a range from 67.5 to 82.1. It suggests a careful tuning of $\alpha$ would make our models achieve desired performance. 2) BGCN-T outperforms vanilla GCN in most cases. It again verifies that the BA is capable of capturing the complex patterns of information propagation, which are hard to reveal by the linear aggregator individually. 3) Surprisingly, the performance of BGCN-T with $\alpha=1$ is much worse than the performance when $\alpha$ is set to the optimal value. One possible reason is that the BA mainly serves as the complementary component to the linear aggregator, hardly working alone to achieve the comparable performance.

\paragraph{Impacts of Multi-Hop Neighbors.}
We also study the effects of $\beta$, in order to explore such trade-off between the aggregated information from different hops. 
Note that setting $\beta$ as 0 and 1 denotes the individual modeling of one- and two-hop neighbors, respectively.
As Figure~\ref{fig:perf_alpha_beta} shows, we observe that involving the pairwise interactions from one- and two-hop neighbors simultaneously achieves better performance. It again verifies the effectiveness of stacking more BAs. 

\subsection{In-Depth Analysis of Aggregators}
We perform in-depth analysis of different aggregators to clarify their working mechanism with respect to two node characteristics ---
1)~\textbf{Degree}, which denotes the average numbers of (one- and two-hop) neighbors surrounding the target node,
and 2)~\textbf{Ratio}, we first count the number of (one- and two-hop) neighbors which have the same label with the target node, and then divide this number by the number of all one- and two-hop neighbors.
We summarize our statistical results in Table~\ref{tab:statistic of nodes}, wherein the symbol $\surd$ and $\times$ denote whether the target nodes are correctly classified or misclassified, respectively. That is, we categorize the testing nodes into three groups according to the correctness of predictions from GCN and BGCN-T. 
Jointly analyzing the three categories corresponding to the three rows in Table~\ref{tab:statistic of nodes}, we have the following findings: 1) Focusing on the third category with the least degree, BGCN-T consistently outperforms GCN, suggesting that the bilinear aggregator is able to distill useful information from sparser neighbors. 2) Comparing the third category to the second one, we observe that BGCN-T is able to endow the predictor node denoising ability. That is, BGCN-T can effectively aggregate information from the neighbors with consistent labels, filtering out the useless information from the irrelevant neighbors. 3) We also realize the limitations of BGCN-T from the second category --- the bilinear interaction might need more label-consistent neighbors (\ie larger ratio), when more neighbors are involved (\ie larger degree).

\begin{table}[t]
	\centering
	\resizebox{0.42\textwidth}{!}{%
	\begin{tabular}{cc|cccccc}
		\hline
		\multirow{2}{*}{GCN}&\multirow{2}{*}{BGCN-T} &\multicolumn{2}{c}{Pubmed} &\multicolumn{2}{c}{Cora} &\multicolumn{2}{c}{Citeseer} \\
		\cline{3-8}
		& &Degree&Ratio &Degree&Ratio &Degree&Ratio \\
		\hline \hline
		$\surd$&$\surd$ &$63.7$ &$0.83$ &$37.8$ &$0.75$ &$16.4$ &$0.77$\\
		$\surd$&$\times$ &$61.2$ &$0.65$ &$32.3$ &$0.77$ &$7.8$ &$0.77$\\
		$\times$&$\surd$ &$45.7$ &$0.59$ &$28.5$ &$0.76$ &$7.6$ &$0.71$\\
		\hline
	\end{tabular}
	}
	\caption{Analysis of aggregators on the test set.}
	\label{tab:statistic of nodes}
\end{table}
\section{Conclusion}
In this paper, we proposed BGNN, a new graph neural network framework, which augments the expressiveness of vanilla GNN by considering the interactions between neighbor nodes. 
The neighbor node interactions are captured by a simple but carefully devised bilinear aggregator. 
The simpleness of the bilinear aggregator makes BGNN have the same model complexity as vanilla GNN \wrt the number of learnable parameters and analytical time complexity. 
Furthermore, the bilinear aggregator is proved to be permutation invariant which is an important property for GNN aggregators~\cite{hamilton2017inductive,xu2018powerful}.
We applied the proposed BGNN on the semi-supervised node classification task, achieving state-of-the-art performance on three benchmark datasets. 
In future, we plan to explore the following research directions: 1) encoding high-order interactions among multiple neighbors, 2) exploring the effectiveness of deeper BGNNs with more than two layers, and 3) developing AutoML technique~\cite{feurer2015efficient} to adaptively learn the optimal $\alpha$ and $\beta$ for each neighbor.
\bibliographystyle{named}
\bibliography{refers}

\begin{thebibliography}{}

\bibitem[\protect\citeauthoryear{Atwood and
  Towsley}{2016}]{atwood2016diffusion}
James Atwood and Don Towsley.
\newblock Diffusion-convolutional neural networks.
\newblock In {\em NeurIPS}, pages 1993--2001, 2016.

\bibitem[\protect\citeauthoryear{Beutel \bgroup \em et al.\egroup
  }{2018}]{beutel2018latent}
Alex Beutel, Paul Covington, Sagar Jain, Can Xu, Jia Li, Vince Gatto, and Ed~H
  Chi.
\newblock Latent cross: Making use of context in recurrent recommender systems.
\newblock In {\em WSDM}, pages 46--54, 2018.

\bibitem[\protect\citeauthoryear{Bronstein \bgroup \em et al.\egroup
  }{2017}]{bronstein2017geometric}
Michael~M Bronstein, Joan Bruna, Yann LeCun, Arthur Szlam, and Pierre
  Vandergheynst.
\newblock Geometric deep learning: going beyond euclidean data.
\newblock {\em IEEE Signal Processing Mag}, 34(4):18--42, 2017.

\bibitem[\protect\citeauthoryear{Bruna \bgroup \em et al.\egroup
  }{2014}]{bruna2013spectral}
Joan Bruna, Wojciech Zaremba, Arthur Szlam, and Yann LeCun.
\newblock Spectral networks and locally connected networks on graphs.
\newblock {\em ICLR}, 2014.

\bibitem[\protect\citeauthoryear{Chami \bgroup \em et al.\egroup
  }{2019}]{chami2019hyperbolic}
Ines Chami, Zhitao Ying, Christopher R{\'e}, and Jure Leskovec.
\newblock Hyperbolic graph convolutional neural networks.
\newblock In {\em NeurIPS}, pages 4869--4880, 2019.

\bibitem[\protect\citeauthoryear{Chen \bgroup \em et al.\egroup
  }{2018}]{chen2018fastgcn}
Jie Chen, Tengfei Ma, and Cao Xiao.
\newblock Fastgcn: fast learning with graph convolutional networks via
  importance sampling.
\newblock {\em ICLR}, 2018.

\bibitem[\protect\citeauthoryear{Defferrard \bgroup \em et al.\egroup
  }{2016}]{defferrard2016convolutional}
Micha{\"e}l Defferrard, Xavier Bresson, and Pierre Vandergheynst.
\newblock Convolutional neural networks on graphs with fast localized spectral
  filtering.
\newblock In {\em NeurIPS}, pages 3844--3852, 2016.

\bibitem[\protect\citeauthoryear{Feng \bgroup \em et al.\egroup
  }{2019}]{feng2019temporal}
Fuli Feng, Xiangnan He, Xiang Wang, Cheng Luo, Yiqun Liu, and Tat-Seng Chua.
\newblock Temporal relational ranking for stock prediction.
\newblock {\em ACM Transactions on Information Systems (TOIS)}, 37(2):1--30,
  2019.

\bibitem[\protect\citeauthoryear{Feurer \bgroup \em et al.\egroup
  }{2015}]{feurer2015efficient}
Matthias Feurer, Aaron Klein, Katharina Eggensperger, Jost Springenberg, Manuel
  Blum, and Frank Hutter.
\newblock Efficient and robust automated machine learning.
\newblock In {\em NeurIPS}, pages 2962--2970, 2015.

\bibitem[\protect\citeauthoryear{Hamilton \bgroup \em et al.\egroup
  }{2017}]{hamilton2017inductive}
Will Hamilton, Zhitao Ying, and Jure Leskovec.
\newblock Inductive representation learning on large graphs.
\newblock In {\em NeurIPS}, pages 1024--1034, 2017.

\bibitem[\protect\citeauthoryear{He and Chua}{2017}]{he2017neural}
Xiangnan He and Tat-Seng Chua.
\newblock Neural factorization machines for sparse predictive analytics.
\newblock In {\em SIGIR}, pages 355--364, 2017.

\bibitem[\protect\citeauthoryear{He \bgroup \em et al.\egroup
  }{2016}]{he2016deep}
Kaiming He, Xiangyu Zhang, Shaoqing Ren, and Jian Sun.
\newblock Deep residual learning for image recognition.
\newblock In {\em CVPR}, pages 770--778, 2016.

\bibitem[\protect\citeauthoryear{He \bgroup \em et al.\egroup
  }{2020}]{he2020lightgcn}
Xiangnan He, Kuan Deng, Xiang Wang, Yan Li, Yongdong Zhang, and Meng Wang.
\newblock Lightgcn: Simplifying and powering graph convolution network for
  recommendation.
\newblock In {\em SIGIR}, 2020.

\bibitem[\protect\citeauthoryear{Kampffmeyer \bgroup \em et al.\egroup
  }{2019}]{kampffmeyer2019rethinking}
Michael Kampffmeyer, Yinbo Chen, Xiaodan Liang, Hao Wang, Yujia Zhang, and
  Eric~P Xing.
\newblock Rethinking knowledge graph propagation for zero-shot learning.
\newblock In {\em CVPR}, pages 11487--11496, 2019.

\bibitem[\protect\citeauthoryear{Kipf and Welling}{2017}]{kipf2016semi}
Thomas~N Kipf and Max Welling.
\newblock Semi-supervised classification with graph convolutional networks.
\newblock {\em ICLR}, 2017.

\bibitem[\protect\citeauthoryear{Li \bgroup \em et al.\egroup
  }{2018}]{li2018deeper}
Qimai Li, Zhichao Han, and Xiao-Ming Wu.
\newblock Deeper insights into graph convolutional networks for semi-supervised
  learning.
\newblock In {\em AAAI}, 2018.

\bibitem[\protect\citeauthoryear{Liao \bgroup \em et al.\egroup
  }{2019}]{liao2018lanczosnet}
Renjie Liao, Zhizhen Zhao, Raquel Urtasun, and Richard Zemel.
\newblock Lanczosnet: Multi-scale deep graph convolutional networks.
\newblock In {\em ICLR}, 2019.

\bibitem[\protect\citeauthoryear{Park and Neville}{2019}]{park2019exploiting}
Hogun Park and Jennifer Neville.
\newblock Exploiting interaction links for node classification with deep graph
  neural networks.
\newblock In {\em IJCAI}, pages 3223--3230, 2019.

\bibitem[\protect\citeauthoryear{Perozzi \bgroup \em et al.\egroup
  }{2014}]{perozzi2014deepwalk}
Bryan Perozzi, Rami Al-Rfou, and Steven Skiena.
\newblock Deepwalk: Online learning of social representations.
\newblock In {\em KDD}, pages 701--710, 2014.

\bibitem[\protect\citeauthoryear{Rendle}{2010}]{rendle2010factorization}
Steffen Rendle.
\newblock Factorization machines.
\newblock In {\em ICDM}, pages 995--1000, 2010.

\bibitem[\protect\citeauthoryear{Sen \bgroup \em et al.\egroup
  }{2008}]{sen2008collective}
Prithviraj Sen, Galileo Namata, Mustafa Bilgic, Lise Getoor, Brian Galligher,
  and Tina Eliassi-Rad.
\newblock Collective classification in network data.
\newblock {\em AI magazine}, 29(3):93--93, 2008.

\bibitem[\protect\citeauthoryear{Veli{\v{c}}kovi{\'c} \bgroup \em et al.\egroup
  }{2018}]{velivckovic2017graph}
Petar Veli{\v{c}}kovi{\'c}, Guillem Cucurull, Arantxa Casanova, Adriana Romero,
  Pietro Lio, and Yoshua Bengio.
\newblock Graph attention networks.
\newblock {\em ICLR}, 2018.

\bibitem[\protect\citeauthoryear{Veličković \bgroup \em et al.\egroup
  }{2019}]{velickovic2018deep}
Petar Veličković, William Fedus, William~L. Hamilton, Pietro Liò, Yoshua
  Bengio, and R~Devon Hjelm.
\newblock Deep graph infomax.
\newblock In {\em ICLR}, 2019.

\bibitem[\protect\citeauthoryear{Wang \bgroup \em et al.\egroup
  }{2019}]{wang2019neural}
Xiang Wang, Xiangnan He, Meng Wang, Fuli Feng, and Tat{-}Seng Chua.
\newblock Neural graph collaborative filtering.
\newblock In {\em SIGIR}, pages 165--174, 2019.

\bibitem[\protect\citeauthoryear{Weston \bgroup \em et al.\egroup
  }{2012}]{weston2012deep}
Jason Weston, Fr{\'e}d{\'e}ric Ratle, Hossein Mobahi, and Ronan Collobert.
\newblock Deep learning via semi-supervised embedding.
\newblock In {\em Neural Networks: Tricks of the Trade}, pages 639--655.
  Springer, 2012.

\bibitem[\protect\citeauthoryear{Wu \bgroup \em et al.\egroup
  }{2019}]{wu2019simplifying}
Felix Wu, Tianyi Zhang, Amauri Holanda~de Souza~Jr, Christopher Fifty, Tao Yu,
  and Kilian~Q Weinberger.
\newblock Simplifying graph convolutional networks.
\newblock {\em ICML}, pages 6861--6871, 2019.

\bibitem[\protect\citeauthoryear{Xinyi and Chen}{2019}]{xinyi2018capsule}
Zhang Xinyi and Lihui Chen.
\newblock Capsule graph neural network.
\newblock In {\em ICLR}, 2019.

\bibitem[\protect\citeauthoryear{Xu \bgroup \em et al.\egroup
  }{2018}]{xu2018representation}
Keyulu Xu, Chengtao Li, Yonglong Tian, Tomohiro Sonobe, Ken-ichi Kawarabayashi,
  and Stefanie Jegelka.
\newblock Representation learning on graphs with jumping knowledge networks.
\newblock {\em ICML}, pages 8676--8685, 2018.

\bibitem[\protect\citeauthoryear{Xu \bgroup \em et al.\egroup
  }{2019a}]{xu2018graph}
Bingbing Xu, Huawei Shen, Qi~Cao, Yunqi Qiu, and Xueqi Cheng.
\newblock Graph wavelet neural network.
\newblock In {\em ICLR}, 2019.

\bibitem[\protect\citeauthoryear{Xu \bgroup \em et al.\egroup
  }{2019b}]{xu2018powerful}
Keyulu Xu, Weihua Hu, Jure Leskovec, and Stefanie Jegelka.
\newblock How powerful are graph neural networks?
\newblock {\em ICLR}, 2019.

\bibitem[\protect\citeauthoryear{Yang \bgroup \em et al.\egroup
  }{2016}]{yang2016revisiting}
Zhilin Yang, William~W Cohen, and Ruslan Salakhutdinov.
\newblock Revisiting semi-supervised learning with graph embeddings.
\newblock {\em ICML}, pages 86--94, 2016.

\bibitem[\protect\citeauthoryear{Zhang \bgroup \em et al.\egroup
  }{2018}]{zhang2018deep}
Ziwei Zhang, Peng Cui, and Wenwu Zhu.
\newblock Deep learning on graphs: A survey.
\newblock {\em arXiv preprint arXiv:1812.04202}, 2018.

\end{thebibliography}

\end{document}